# Low Size-Complexity Inductive Logic Programming: The East-West Challenge Considered as a Problem in Cost-Sensitive Classification

Peter Turney
Institute for Information Technology
National Research Council Canada
Ottawa, Ontario, Canada
K1A 0R6
peter@ai.iit.nrc.ca

**Abstract**

The Inductive Logic Programming community has considered proof-complexity and model-complexity, but, until recently, size-complexity has received little attention. Recently a challenge was issued "to the international computing community" to discover low size-complexity Prolog programs for classifying trains. The challenge was based on a problem first proposed by Ryszard Michalski, 20 years ago. We interpreted the challenge as a problem in cost-sensitive classification and we applied a recently developed cost-sensitive classifier to the competition. Our algorithm was relatively successful (we won a prize). This paper presents our algorithm and analyzes the results of the competition.

## 1. Introduction

Inductive Logic Programming (ILP) has been defined as the intersection of Induction and Logic Programming (Lavrac & Dzeroski, 1994). The problem of induction is often portrayed as the problem of discovering a simple theory for a given set of data. One difficulty with this view is that there are many ways to measure the simplicity (equivalently, the complexity) of a theory. The ILP community has considered proof-complexity (Muggleton *et al*., 1992) and model-complexity (Conklin & Witten, 1994), but, until recently, size-complexity has received little attention.[1] Muggleton, Srinivasan, and Bain (1992) presented an algorithm for inducing logic programs with simple proof trees (low proof-complexity). Conklin and Witten (1994) presented an algorithm for inducing logic programs with simple models (low model-complexity). This paper introduces an algorithm for inducing small logic programs (low size-complexity).

Michie *et al.* (1994) issued a challenge "to the international computing community" to discover low size-complexity Prolog programs for classifying trains.[2] The challenge was inspired by a problem posed by Ryszard Michalski (Michalski & Larson, 1977), illustrated in Figure 1. The problem is to discover a simple rule that distinguishes Eastbound trains from Westbound trains. An example is, "If a train has a short closed car, then it is Eastbound and otherwise Westbound." Michie *et al.* (1994) created three separate competitions, based on this problem.

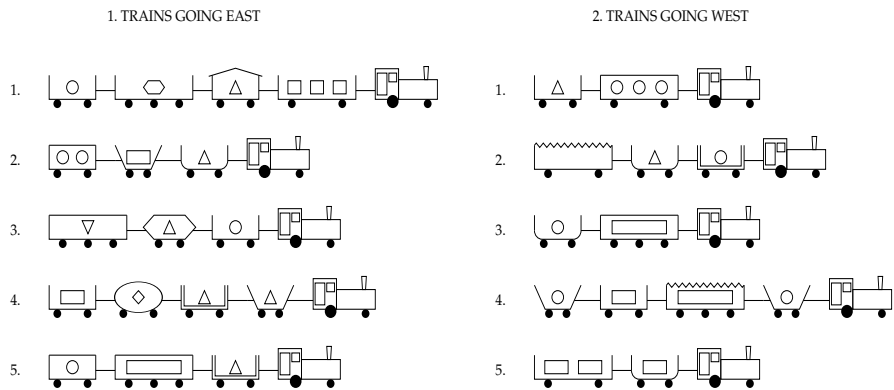

Figure 1. Michalski's original set of 10 trains (Michie *et al.,* 1994).

The first competition involved extending Michalski's set of 10 trains with 10 more trains, generated by a Prolog program written by Stephen Muggleton. The 10 new trains are shown in Figure 2. The combined set of 20 trains was classified into East and West using an arbitrary human-generated rule, called "Theory X". The challenge was to discover Theory X or a simpler theory (rule). The complexity of a rule was measured by representing the rule as a Prolog program and calculating the sum of the number of clause occurrences, the number of term occurrences, and the number of atom occurrences.

The second competition was intended for learning algorithms that cannot easily represent in Prolog what they have learned. For example, a neural network could learn to classify trains, but it would be difficult to convert the network into a simple Prolog repre-sentation. Although the second competition was intended for sub-symbolic and semi-symbolic inductive analysis, symbolic learners were not forbidden. The second competition involved classifying 100 new trains, using a classifier that had been trained on the 20 trains of Figures 2 and 2. The oracle for correct classification was Theory X or any theory that scored in the bottom quartile (the simplest 25%) of the theories in the first competition. In this competition, theories were evaluated by their behavior, rather than by their size-complexity. However, the standard for correct behavior was defined by the simpler (smaller) theories of the first competition. In our analysis of the results of the second competition, we focus on the human-generated Theory X as an oracle, rather than the machine-generated bottom quartile theories.

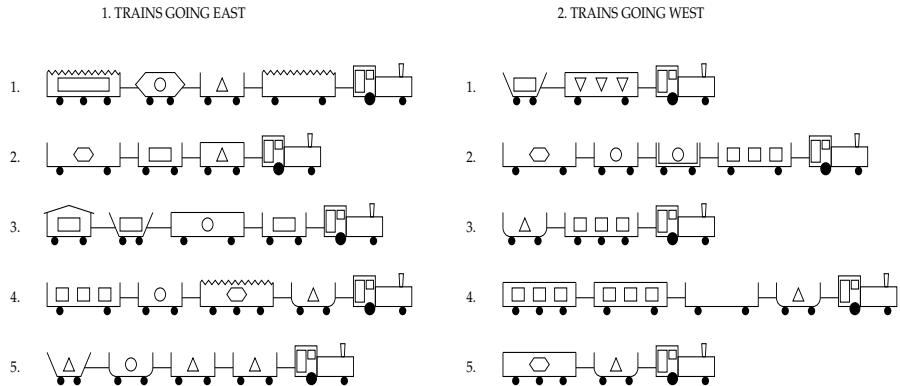

Figure 2. The new set of 10 trains (Michie *et al.,* 1994).

The third competition involved classifying five sets of ten trains. Each set of ten was randomly split into five Eastbound trains and five Westbound trains. The challenge was to generate five theories, one for each set of trains, with minimal total complexity. Complexity was measured as in the first competition, by using the Prolog representation of the theories.

In the East-West Challenge, the input data (the descriptions of the trains) and the output theories (in the first and third competitions) are represented in Prolog. If ILP is defined as the intersection of Induction and Logic Programming (Lavrac & Dzeroski, 1994), then, by definition, any computer-generated or computer-assisted entries in the East-West Challenge were the product of ILP. The challenge placed no constraints on how the theories could be generated. Most of the entries in the first competition were human-generated, without computer assistance, but the better entries were computer-generated or computer-assisted. Bernhard Pfahringer of the Austrian Research Institute for Artificial Intelligence used a clever exhaustive search algorithm to win the first and third competitions.[3] The algorithm described here won the second competition.[4]

We recently developed a cost-sensitive algorithm, called ICET, for generating low-cost decision trees (Turney, 1995). ICET takes feature vectors as input and generates decision trees as output. The algorithm is sensitive to both the cost of the features (attributes, measurements, tests) and the cost of classification errors. For the East-West Challenge, we extended ICET to handle Prolog input. The decision tree output was converted to Prolog manually.[5] We call this algorithm RL-ICET (Relational Learning with ICET).

RL-ICET is similar to the LINUS (Lavrac & Dzeroski, 1994) ILP system. RL-ICET uses a three-part strategy: First, a *pre-processor* translates the Prolog relations and predicates into a feature vector format. The pre-processor in RL-ICET was designed specially for the East-West Challenge, although it could be generalized to other domains. LINUS (Lavrac & Dzeroski, 1994) has a general-purpose pre-processor. Second, an *attribute-value learner* applies a decision tree induction algorithm (ICET) to the feature vectors. Each feature is assigned a cost, based on the size of the fragment of Prolog code that represents the corresponding predicate or relation. A decision tree that has a low cost corresponds (roughly) to a Prolog program that has a low size-complexity. When it searches for a low cost decision tree, ICET is in effect searching for a low size-complexity Prolog program. The attribute-value learners in LINUS are not cost-sensitive. Third, a *post-processor* translates the decision tree into a Prolog program. Post-processing with RL-ICET was done manually. LINUS performs post-processing automatically.

ILP algorithms may be characterized by how they search the hypothesis space. ILP systems typically use operations such as *inverse resolution* (Muggleton & Buntine, 1988), *relative least general generalization* (Muggleton & Feng, 1990), or *top-down search of refinement graphs* (Shapiro, 1983). Like LINUS, RL-ICET can be viewed as a form of top-down search of refinement graphs.

We describe ICET in depth in Section 2. We then explain how ICET was extended to RL-ICET, for the East-West Challenge, in Section 3. In Section 4, we discuss the results of the East-West Challenge. Although we won a prize in the second competition, Bernhard Pfahringer's exhaustive search algorithm is clearly the best strategy for this type of competition. However, it is not clear that Pfahringer's algorithm has any applications outside this type of competition. On the other hand, ICET has been successfully applied to five real-world medical datasets (Turney, 1995). We conclude that the East-West Challenge provides further evidence of the versatility of the ICET algorithm and the power of the LINUS approach to ILP.

## 2. The ICET Algorithm

ICET is a hybrid of a genetic algorithm and a decision tree induction algorithm (Turney, 1995). The genetic algorithm is Grefenstette's (1986) GENESIS and the decision tree induction algorithm is Quinlan's (1993) C4.5. ICET uses a two-tiered search strategy. On the bottom tier, C4.5 uses a TDIDT (Top Down Induction of Decision Trees) strategy to search through the space of decision trees. On the top tier, GENESIS uses a genetic algorithm to search through the space of biases.

ICET accepts as input a set of training data in the form of feature vectors. Suppose there are $n$ features (i.e., tests, attributes, variables, fields) in the data. The bias $\beta$ is specified by a vector of $n + 2$ numbers:

$$\beta = \langle B_1, ..., B_n, \omega, \text{CF} \rangle \tag{1}$$

$$0 \leq B_i \leq 10000 \tag{2}$$

$$0 \leq \omega \leq 1 \tag{3}$$

$$1 \leq \text{CF} \leq 100 \tag{4}$$

We have modified C4.5 so that its bias can be adjusted by these $n + 2$ numbers. At each step in the construction of a decision tree, C4.5 chooses the attribute that maximizes the following formula (Núñez, 1988, 1991):

$$\frac{2^{\Delta I_i} - 1}{(B_i + 1)^{\omega}} \tag{5}$$

In this formula, $\Delta I_i$ is the information gain associated with the $i$-th attribute at a given stage in the construction of the decision tree. $B_i$ is the bias associated with the $i$-th attribute. The original C4.5 selects the attribute that maximizes $\Delta I_i$.[6] Our modified C4.5 uses $B_1, ..., B_n$ as adjustable bias parameters that make C4.5 prefer one attribute over another. The modified C4.5 tends to avoid the $i$-th attribute when $B_i$ is large. The parameter $\omega$ adjusts the strength of avoidance. The bias parameters $B_1, ..., B_n$ are ignored when $\omega = 0$ and they have maximum influence when $\omega = 1$. The parameter CF (part of the original C4.5) is used to control the level of pruning of the decision tree.

The GENESIS algorithm was inspired by biological evolution, in which a population of individuals evolves over successive generations (Grefenstette, 1986). In ICET, a population consists of 50 bias vectors $\beta_1, ..., \beta_{50}$, which evolve over 20 generations.[7] The fitness of an individual $\beta_i$ is determined by applying C4.5 to the given data, using the values of $\langle B_1, ..., B_n, \omega, \text{CF} \rangle$ that are specified by $\beta_i$. The fitness is the average cost of using the decision tree that is generated by C4.5. The cost includes both the costs of the features and the cost of misclassification errors.

The population in the first generation is randomly generated. In subsequent generations, the population is generated by applying mutation and crossover (i.e., mating pairs of individuals) to the individuals in the previous generation. Fitter individuals have more offspring in the next generation. The output of ICET is the fittest decision tree, over the 20 generations. This algorithm is sketched in Figure 3 (the population is shown as three individuals, instead of 50, to make the figure intelligible).

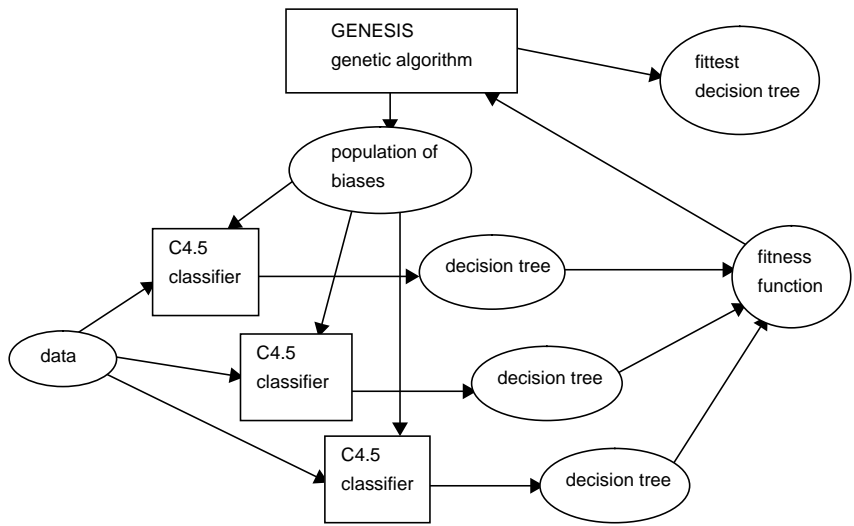

Figure 3. A sketch of the ICET algorithm (Turney, 1995).

The ICET algorithm is discussed in detail elsewhere (Turney, 1995), including ICET's relationship with other cost-sensitive algorithms and other two-tiered search strategies. ICET was empirically evaluated using five real-world medical datasets. It was found to be superior to three other cost-sensitive algorithms — EG2 (Núñez, 1988, 1991), IDX (Norton, 1989), and CS-ID3 (Tan & Schlimmer, 1989, 1990; Tan, 1993) — and to C4.5 (Quinlan, 1993), which ignores cost.

ICET assumes that the data are in the form of feature vectors and it generates output in the form of decision trees. The East-West Challenge involves data in the form of predicates and relations and theories in the form of Prolog programs. The next section explains how we handled this difficulty.

## 3. ICET Applied to the East-West Challenge

The raw data in the East-West challenge were represented using Prolog. For example, this is the representation of the first train in Figure 1:

    eastbound([c(1, rectangle, short, not_double, none, 2, l(circle, 1)),
        c(2, rectangle, long, not_double, none, 3, l(hexagon, 1)),
        c(3, rectangle, short, not_double, peaked, 2, l(triangle, 1)),
        c(4, rectangle, long, not_double, none, 2, l(rectangle, 3))]).

We wrote a simple Prolog program that converted the data into the feature vector (attribute-value) format that is typically used in decision tree induction.

We started with 28 predicates that apply to the cars in a train, such as ellipse(C), which is true when the car C has an elliptical shape. For each of these 28 predicates, we defined a corresponding feature. All of the features were defined for whole trains, rather than single cars, since the problem is to classify trains. The feature ellipse, for example, has the value 1 when a given train has a car with an elliptical shape. Otherwise ellipse has the value 0. We then defined $(28 \times 27) / 2 = 378$ features by forming all possible unordered pairs of the original 28 predicates. For example, the feature ellipse_triangle_load has the value 1 when a given train has a car with an elliptical shape that is carrying a triangle load, and 0 otherwise.[8] Note that the features ellipse and triangle_load may have the value 1 for a given train while the feature ellipse_triangle_load has the value 0, since ellipse_triangle_load only has the value 1 when the train has a car that is *both* elliptical and carrying a triangle load. We then defined $28 \times 28 = 784$ features by forming all possible ordered pairs of the original 28 predicates, using the relation infront(T, C1, C2). For example, u_shaped_infront_peaked_roof has the value 1 when the train has a U-shaped car in front of a car with a peaked roof, and 0 otherwise. Finally, we added 9 more predicates that apply to the train as a whole, such as train_4, which has the value 1 when the train has exactly four cars. Thus a train is represented by a feature vector with $28 + 378 + 784 + 9 = 1199$ elements, where every feature has either the value 1 or the value 0. (The 28 car predicates and 9 train predicates are listed in the Appendix.)

The feature vector for a train does not capture all the information that is in the original Prolog representation. For example, we could also define $(28 \times 27 \times 26) / (3 \times 2) = 3276$ features by combining all possible unordered triples of the 28 predicates. However, these features would likely be less useful, since they are so specific that they will only rarely have the value 1. If the target concept should happen to be a triple of predicates, it could be closely approximated by the conjunction of the three pairs of predicates that are subsets of the triple.

This kind of translation to feature vector representation could be applied to many other types of structured objects. For example, consider the problem of classifying a set of documents. The keywords in a document are analogous to the cars in a train. The distance between keywords or the order of keywords in a document may be useful when classifying the document, just as the infront relation may be useful when classifying trains.

LINUS (Lavrac & Dzeroski, 1994) also has a pre-processor (called the *DHDB interface*) that translates relational descriptions into feature vector format. The DHDB interface consists of more than 2000 lines of Prolog code. It is much more general purpose than the little Prolog program we used here, which is tailored specially for the East-West Challenge.

Each feature was assigned a cost, based on the complexity of the fragment of Prolog code required to represent the given feature. Recall that the complexity of a Prolog program is defined as the number of clause occurrences plus the number of term occurrences plus the number of atom occurrences.[9] Table 1 shows some examples of features and their costs.

Table 1. Examples of features and their costs.

| Feature | Prolog Fragment | Cost (Complexity) |
| --- | --- | --- |
| ellipse | has_car(T, C), ellipse(C). | 5 |
| short_closed | has_car(T, C), short(C), closed(C). | 7 |
| train_4 | len1(T, 4). | 3 |
| train_hexagon | has_load1(T, hexagon). | 3 |
| ellipse_peaked_roof | has_car(T, C), ellipse(C), arg(5, C, peaked). | 9 |
| u_shaped_no_load | has_car(T, C), u_shaped(C), has_load(C, 0). | 8 |
| rectangle_load_infront_jagged_roof | infront(T, C1, C2), has_load0(C1, rectangle), arg(5, C2, jagged). | 11 |

In addition to the cost of the features, ICET takes as input the cost of classification errors. The rules of the contest required that the theories perform perfectly on the training data (there was no independent testing dataset), so we arbitrarily assigned a cost of 1000 to classification errors. This penalty was high enough to ensure that the genetic algorithm would eliminate a bias that resulted in a decision tree that made mistakes. The fitness of a decision tree was calculated by applying the decision tree to the training data. For example, in the first competition, the training data consisted of the 20 trains in Figures 1 and 2, represented as feature vectors. The fitness of a decision tree was the cost of using the tree on the training data, which was the sum of the costs of all of the tests in the tree and the error rate of the tree multiplied by 1000 (the cost of an error).

Since the initial population of biases in ICET is set randomly, ICET may produce a different result each time it runs. We ran ICET several times. Figure 4 shows the best (lowest cost) decision tree that was generated for the first competition. Each run took about four to five hours on a single-processor Sun Sparc 10. The cost of the best decision tree in the final generation (the 20th generation) was typically about half the cost of the best decision tree in the first generation.

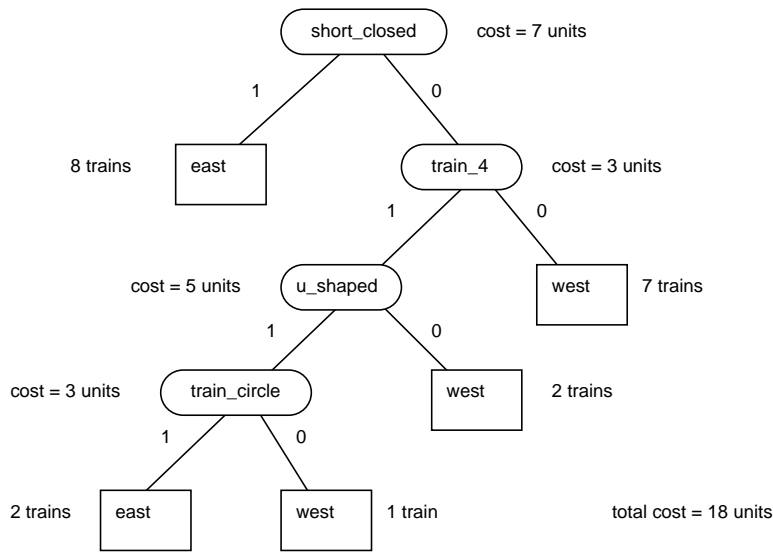

Figure 4. The best decision tree for the first competition.

LINUS (Lavrac & Dzeroski, 1994) also has an attribute-value learning component. This component consists of three attribute-value learners — NEWGEM (Mozetic, 1985), ASSISTANT (Cestnik *et al.*, 1987), and CN2 (Clark & Boswell, 1991). Unlike ICET, none of the three attribute-value learners is cost-sensitive. However, each learner has a different bias, so LINUS is performing a limited search in bias space by using three different attribute-value learners.[10]

We converted the decision trees into Prolog programs by hand. For example, Figure 4 was converted to the following Prolog program:

```
eastbound(T) :-
        has_car(T, C),
        ((short(C), closed(C)) ;
        (len1(T, 4), u_shaped(C), has_load1(T, circle))).
```

The above Prolog program was our entry for the first competition.

This program has a complexity of 19, which shows that the cost of the decision tree (18, as we can see in Figure 4) is only an approximation of the cost of the corresponding Prolog program. We need to add some Prolog code to assemble the Prolog fragments into a working whole. This extra code means that the sum of the sizes of the fragments is less than the size of the whole program. It is also sometimes possible to subtract some code from the whole, because there may be some overlap in the code in the fragments. The ideal solution to this problem would be to add a post-processing module to RL-ICET that automatically converts the decision trees into Prolog programs. The complexity could then be calculated directly from the output Prolog program, instead of the decision tree. Although post-processing with RL-ICET was done manually, it could be automated, as demonstrated by LINUS (Lavrac & Dzeroski, 1994), which has a general-purpose post-processor.

Other researchers have applied genetic algorithms to ILP problems. Wong and Leung (1994) compared their GLPS (Genetic Logic Programming System) to Quinlan's (1990, 1991) FOIL system on a noisy version of the chess end-game problem and they found GLPS to be superior to FOIL. They also successfully applied GLPS to the network reachability problem and the factorial problem. Wong and Leung's (1994) GLPS system is similar to our approach to the East-West Challenge only in that both use a genetic algorithm. GLPS operates directly on Prolog programs and does not involve decision trees in any way. It would be interesting to see how well GLPS performs on the East-West Challenge.

## 4. Analysis of the Results of the Competition

In this section, we present our interpretation of the results of the three competitions in the East-West Challenge.[3]

There were 63 entries in the first competition. After several entries were dropped, because they were too complex or they were inconsistent, 39 entries were left. The winner was Bernhard Pfahringer's entry, called pfahr2. The algorithm he used was an efficient form of exhaustive search through the space of Prolog programs. The Prolog program pfahr2 has a size-complexity of 16. Four theories tied for second place, with a size-complexity of 19. Table 2 shows the results for the top seven entries, where complexity ranged from 16 to 20. Theory X (the "true" theory, that was used to classify the 20 trains in competition 1) has a size of 19. The remaining 32 theories had sizes ranging from 22 to 43.

Table 2. Results for the seven best entries in competition 1.

| Entry Name | Size | Agreement with pfahr2 | Agreement with X | Contestant |
| --- | --- | --- | --- | --- |
| pfahr2 | 16 | 100% | 80% | Bernhard Pfahringer |
| inglis | 19 | 56% | 50% | Stuart Inglis |
| pfahr1 | 19 | 56% | 50% | Bernhard Pfahringer |
| turney | 19 | 72% | 90% | Peter D. Turney |
| weka | 19 | 56% | 50% | WEKA ML Project |
| aqdt1 | 20 | 61% | 47% | Ibrahim F. Imam |
| aqdt2 | 20 | 60% | 52% | Ryszard S. Michalski |

In addition to the size of the theories, Table 2 shows the agreement between each theory and the winning theory (pfahr2) and the "true" theory (Theory X). We define *agreement* using the idea of the second competition. We applied each of the theories to the 100 trains used in the second competition. The agreement between two theories was measured by the percentage of the 100 trains for which the two theories agreed on the classification (East or West).[11] Our entry (turney) has the highest agreement with Theory X and the highest agreement with pfahr2 (excluding the agreement of pfahr2 with itself). None of the entries exactly matched Theory X. Theory X and pfahr2 were both recursive Prolog programs. Due to its design, our algorithm is incapable of generating recursive theories.[12]

There were seven entries in the second competition. The winner was our entry (turney). The second competition was intended for sub-symbolic learners; learners that cannot readily express their theories as Prolog programs. The idea in the second competition was to score the theories purely by their behavior. An entry in the competition consisted of an assignment of the labels East and West to a set of 100 trains. The standard for correct classification could be any theory in the bottom quartile of theories in the first competition. There was a loophole in the second competition, since there was no rule that forbids submitting the same entry to both the first and second competitions. Three entries exploited this loophole (turney, aqdt1, and aqdt2) and all three theories scored 100%, each theory using itself as the standard for correct classification. The tie was broken by using pfahr2 as the standard.

We need not stick to the original rules of the second competition in our analysis of the results. It seems that it would be more fair to judge the entries by their agreement with either pfahr2 or Theory X. Table 3 uses these two oracles to score the entries for the second competition. Our entry is still the most successful.

Table 3. Results for competition 2.

| Entry Name | Agreement with pfahr2 | Agreement with X | Contestant |
| --- | --- | --- | --- |
| aq17hci | 58% | 52% | Janusz Wnek |
| aqdt1 | 61% | 47% | Ibrahim F. Imam |
| aqdt2 | 60% | 52% | Ryszard S. Michalski |
| hart | 61% | 58% | G. R. Hart |
| imam | 60% | 58% | Ibrahim F. Imam |
| quin | 62% | 53% | Ross Quinlan |
| turney | 72% | 90% | Peter D. Turney |

There were six entries in the third competition. Three entries were disqualified for various reasons. The winner was Bernhard Pfahringer. The sum of the sizes of his five theories for the five sets of trains was 74. One of our five theories was disqualified, because we used the Quintus Prolog definition of append, instead of the version of append that was included in the rules of the competition. Table 4 shows the results for the third competition, including the disqualified entries. If we assume that the disquali-fied entries can be repaired, Bernhard Pfahringer is still the winner. Our entry (turney) would then have the second lowest complexity.

Table 4. Results for competition 3, including disqualified entries.

| Entry Name | Total of Sizes | Qualified | Contestant |
|---|---|---|---|
| pfahr | 74 | yes | Bernhard Pfahringer |
| turney | 82 | no | Peter D. Turney |
| aqdt | 85 | no | Ibrahim F. Imam |
| dm2 | 114 | yes | Donald Michie |
| royc | 122 | no | John Roycroft |
| mli | 158 | yes | Mark Maloof |

## 5. Conclusions

To summarize the results of the East-West Challenge, it seems reasonable to say that Bernhard Pfahringer's algorithm is the best and our algorithm is the second best. Pfahringer's algorithm is certainly the best strategy for a competition of this kind, but it seems unlikely that the algorithm can be applied to any real-world problems, due to scaling problems. The size of the search space grows exponentially as the size of the programs increases, so an exhaustive search quickly becomes infeasible. On the other hand, ICET has been successfully applied to five real-world medical datasets (Turney, 1995). The performance of our algorithm in the East-West Challenge (1) indicates the versatility of the ICET algorithm, (2) provides evidence in support of the value of the LINUS (Lavrac & Dzeroski, 1994) approach to ILP, and (3) provides evidence in support of the value of genetic algorithms applied to ILP (Wong & Leung, 1994).

## Acknowledgments

Thanks to Ivan Bruha for pointing out the connection between this work and the LINUS system (Lavrac & Dzeroski, 1994). Thanks to Joel Martin, Grigoris Karakoulas, and three anonymous referees of ILP-95 for their helpful comments on earlier versions of this paper.

**Notes**

1. Although many ILP algorithms, such as FOIL (Quinlan, 1990, 1991), have a bias in favor of low size-complexity, they do not attempt to systematically search for the lowest size-complexity.
2. Michie *et al.* (1994) is available as a PostScript file, ml-chall.ps, included in URL ftp://ftp.comlab.ox.ac.uk/pub/Packages/ILP/trains.tar.Z.
3. The results of the competition are discussed in URL ftp://ftp.comlab.ox.ac.uk/pub/Packages/ILP/results.tar.Z.
4. There was a loophole in the definition of the second competition. This is discussed in Section 4.
5. There was a certain amount of creativity in this manual conversion to Prolog. The conversion could be done automatically, but there was insufficient time to implement it before the deadline for the competition. Our entries might be better described as "computer-assisted", rather than "computer-generated".
6. The original C4.5 (Quinlan, 1993) allows the user to specify either *information gain* $\Delta I_i$ or *information gain ratio* as the selection criterion. The information gain ratio is a function of $\Delta I_i$ and the number of values that the *i*-th attribute may have.
7. The population size of 50 and the choice of 20 generations are the default settings in GENESIS. We have not experimented with alternative settings.
8. Some of these features are not possible. For example, no car can be both elliptical and rectangular, so the feature ellipse_rectangle always has the value 0. We did not bother to eliminate these impossible features, since the ICET algorithm automatically discards them, because their information gain is zero.
9. This definition omits some subtleties that are discussed in Michie *et al.* (1994).
10. It appears that the LINUS user must manually choose one of the three attribute-value learners, NEWGEM, ASSISTANT, or CN2. This is less powerful than RL-ICET, where the genetic algorithm automatically searches bias space.
11. The agreement information in Table 2 was not supplied in URL ftp://ftp.comlab.ox.ac.uk/pub/Packages/ILP/results.tar.Z. We could only calculate the agreement for the top seven theories, since the file that describes the results only supplies the Prolog code for the top seven theories.
12. LINUS (Lavrac & Dzeroski, 1994) cannot generate recursive theories, but DINUS, an extended version of LINUS, *can* generate recursive theories.

## Appendix

For those who would like to reproduce the results described in this paper, this appendix lists the predicates that were used to define the 1199 features. To understand this list, the reader will also need to see Michie *et al.* (1994).[2] Here are the 28 car predicates that are discussed in Section 3:

```
ellipse(C) :- arg(2, C, ellipse).
hexagon(C) :- arg(2, C, hexagon).
rectangle(C) :- arg(2, C, rectangle).
u_shaped(C) :- arg(2, C, u_shaped).
bucket(C) :- arg(2, C, bucket).
long(C) :- arg(3, C, long).
short(C) :- arg(3, C, short).
double(C) :- arg(4, C, double).
not_double(C) :- not double(C).
open(C) :- arg(5, C, none).
closed(C) :- not open(C).
no_roof(C) :- arg(5, C, none).
flat_roof(C) :- arg(5, C, flat).
jagged_roof(C) :- arg(5, C, jagged).
peaked_roof(C) :- arg(5, C, peaked).
arc_roof(C) :- arg(5, C, arc).
two_axles(C) :- arg(6, C, 2).
three_axles(C) :- arg(6, C, 3).
circle_load(C) :- has_load0(C, circle).
hexagon_load(C) :- has_load0(C, hexagon).
rectangle_load(C) :- has_load0(C, rectangle).
triangle_load(C) :- has_load0(C, triangle).
diamond_load(C) :- has_load0(C, diamond).
utriangle_load(C) :- has_load0(C, utriangle).
no_load(C) :- has_load(C, 0).
one_load(C) :- has_load(C, 1).
two_load(C) :- has_load(C, 2).
three_load(C) :- has_load(C, 3).
```

Here are the 9 train predicates that are discussed in Section 3:

```
train_2(T) :- len1(T, 2).
train_3(T) :- len1(T, 3).
```

```
train_4(T) :- len1(T, 4).
train_circle(T) :- has_load1(T, circle).
train_hexagon(T) :- has_load1(T, hexagon).
train_rectangle(T) :- has_load1(T, rectangle).
train_triangle(T) :- has_load1(T, triangle).
train_diamond(T) :- has_load1(T, diamond).
train_utriangle(T) :- has_load1(T, utriangle).
```

## References

Cestnik, B., Kononenko, I., & Bratko, I. (1987). ASSISTANT 86: A knowledge elicitation tool for sophisticated users. In I. Bratko and N. Lavrac, editors, *Progress in Machine Learning*, pp. 31-45. Wilmslow, UK: Sigma Press.

Clark, P., & Boswell, R. (1991). Rule induction with CN2: Some recent improvements. In *Proceedings of the Fifth European Working Session on Learning, EWSL-91*, pp. 151-163. Berlin: Springer Verlag.

Conklin, D., & Witten, I.H. (1994). Complexity-based induction. *Machine Learning,* 16, 203-225.

Grefenstette, J.J. (1986). Optimization of control parameters for genetic algorithms. *IEEE Transactions on Systems, Man, and Cybernetics*, 16, 122-128.

Lavrac, N., & Dzeroski, S. (1994). *Inductive Logic Programming: Techniques and Applications.* New York: Ellis Horwood.

Michalski, R.S., & Larson, J.B. (1977). Inductive inference of VL decision rules. Paper presented at Workshop in Pattern-Directed Inference Systems, Hawaii, 1977. *SIGART Newsletter*, ACM, 63, 38-44.

Michie, D., Muggleton, S., Page, D., & Srinivasan, A. (1994). To the international computing community: A new East-West challenge. Oxford University Computing Laboratory, Oxford, UK.

Mozetic, I. (1985). NEWGEM: Program for learning from examples, technical documentation and user's guide. Reports of Intelligent Systems Group, UIUCDCS-F-85-949, Department of Computer Science, University of Illinois, Urbana Champaign, Illinois.

Muggleton, S., & Buntine, W. (1988). Machine invention of first-order predicates by inverting resolution. *Proceedings of the Fifth International Conference on Machine Learning, ML-88,* pp. 339-352. California: Morgan Kaufmann.

Muggleton, S., & Feng, C. (1990). Efficient induction of logic programs. *Proceedings of the First Conference on Algorithmic Learning Theory*, pp. 368-381. Ohmsha, Tokyo.

Muggleton, S., Srinivasan, A., & Bain, M. (1992). Compression, significance and accuracy. In D. Sleeman and P. Edwards, editors, *Machine Learning: Proceedings of the Ninth International Conference (ML92)*, pp. 338-347. California: Morgan Kaufmann.

Norton, S.W. (1989). Generating better decision trees. *Proceedings of the Eleventh International Joint Conference on Artificial Intelligence, IJCAI-89,* pp. 800-805. Detroit, Michigan.

Núñez, M. (1988). Economic induction: A case study. *Proceedings of the Third European Working Session on Learning, EWSL-88*, pp. 139-145. California: Morgan Kaufmann.

Núñez, M. (1991). The use of background knowledge in decision tree induction. *Machine Learning*, 6, 231-250.

Quinlan, J.R. (1990). Learning logical definitions from relations. *Machine Learning*, 5, 239-266.

Quinlan, J.R. (1991). Determinate literals in inductive logic programming. *Proceedings of the Eighth International Workshop on Machine Learning, ML-91,* pp. 442-446. California: Morgan Kaufmann.

Quinlan, J.R. (1993). *C4.5: Programs for machine learning*. California: Morgan Kaufmann.

Shapiro, E. (1983). *Algorithmic Program Debugging.* Massachusetts: MIT Press.

Tan, M., & Schlimmer, J. (1989). Cost-sensitive concept learning of sensor use in approach and recognition. *Proceedings of the Sixth International Workshop on Machine Learning, ML-89,* pp. 392-395. Ithaca, New York.

Tan, M., & Schlimmer, J. (1990). CSL: A cost-sensitive learning system for sensing and grasping objects. *IEEE International Conference on Robotics and Automation*. Cincinnati, Ohio.

Tan, M. (1993). Cost-sensitive learning of classification knowledge and its applications in robotics. *Machine Learning,* 13, 7-33.

Turney, P. (1995). Cost-sensitive classification: Empirical evaluation of a hybrid genetic decision tree induction algorithm. *Journal of Artificial Intelligence Research*, 2, 369-409. [Available on the Internet at URL http://www.cs.washington.edu/research/jair/home.html.]

Wong, M.L., & Leung, K.S. (1994). Inductive logic programming using genetic algorithms. *Advances in Artificial Intelligence — Theory and Application II*, Volume II of the *Proceedings of the 7th International Conference on Systems Research, Informatics and Cybernetics*, Baden-Baden, Germany, pp. 119-124.